\documentclass[letterpaper, 10 pt, conference]{ieeeconf}  

\IEEEoverridecommandlockouts                              

\usepackage{amsmath} 
\usepackage{amssymb}  
\usepackage{graphicx}
\usepackage[table]{xcolor}
\usepackage{hyperref}
\usepackage{booktabs}
\usepackage{adjustbox}
\usepackage{stfloats}
\usepackage{multirow}
\usepackage{array}
\newcommand{\ieeetablesetup}{%
  \centering
  \footnotesize
  \setlength{\tabcolsep}{3.5pt}%
  \renewcommand{\arraystretch}{1.15}%
}
\definecolor{darkblue}{rgb}{0.0, 0.0, 0.55}
\hypersetup{
    colorlinks=true,
    linkcolor=darkblue,
    citecolor=darkblue,
    filecolor=magenta,      
    urlcolor=black,
}

\title{\LARGE \bf
Autonomous UAV Flight Navigation in Confined Spaces: A Reinforcement Learning Approach
}

\author{Marco S. Tayar$^{1}$, Lucas K. de Oliveira$^{1}$, Felipe Andrade G. Tommaselli$^{1}$, Juliano D. Negri$^{1}$, \\
Thiago H. Segreto$^{1}$, Ricardo V. Godoy$^{1}$, and Marcelo Becker$^{1}$ 
\thanks{
This work was supported by the Petr\'{o}leo Brasileiro S/A - Petrobras,
using resources from the R\&D clause of the ANP, in partnership with the
Universidade de S\~{a}o Paulo (USP) and the Funda\c{c}\~{a}o de Apoio \`{a} F\'{\i}sica e
\`{a} Qu\'{\i}mica (FAFQ), under Cooperation Agreement No. 2023/00016-6 and
2023/00013-7.}
\thanks{
$^{1}$Marco S. Tayar, Lucas K. de Oliveira, Felipe Andrade G. Tommaselli, Juliano D. Negri, Thiago H. Segreto, Ricardo V. Godoy, and Marcelo Becker are with the  Department of Mechanical Engineering, University of São Paulo, São Carlos, Brazil.
{\tt\small becker@sc.usp.br}}%
}

\begin{document}

\maketitle
\thispagestyle{empty}
\pagestyle{empty}

\begin{abstract}

Autonomous UAV inspection of confined industrial infrastructure, such as ventilation ducts, demands robust navigation policies where collisions are unacceptable. While Deep Reinforcement Learning (DRL) offers a powerful paradigm for developing such policies, it presents a critical trade-off between on-policy and off-policy algorithms. Off-policy methods promise high sample efficiency, a vital trait for minimizing costly and unsafe real-world fine-tuning. In contrast, on-policy methods often exhibit greater training stability, which is essential for reliable convergence in hazard-dense environments. This paper directly investigates this trade-off by comparing a leading on-policy algorithm, Proximal Policy Optimization (PPO), against an off-policy counterpart, Soft Actor-Critic (SAC), for precision flight in procedurally generated ducts within a high-fidelity simulator. Our results show that PPO consistently learned a stable, collision-free policy that completed the entire course. In contrast, SAC failed to find a complete solution, converging to a suboptimal policy that navigated only the initial segments before failure. This work provides evidence that for high-precision, safety-critical navigation tasks, the reliable convergence of a well-established on-policy method can be more decisive than the nominal sample efficiency of an off-policy algorithm.

\end{abstract}

\section{Introduction}

Manual inspection of industrial infrastructure, such as pipelines and ventilation ducts, is a complex, costly, and time-consuming process that is essential for maintaining operational integrity.
Unmanned Aerial Vehicles (UAVs) represent a significant advancement in the field of industrial inspection, enabling automated and safe data collection in environments that are inaccessible or unsafe for humans. Among the most critical, yet common, of these environments are industrial ducts, which represent confined, often hazardous, spaces critical to facility operations. However, navigating a UAV in ducts presents unique challenges. 
In these environments, the close proximity of walls creates complex aerodynamic effects that increase collision risk~\cite{martin2025flying}.

Classical motion planning methods lack the adaptability needed for these challenging spaces, struggling to handle unmodeled aerodynamic phenomena such as the ground effect inside narrow pipes~\cite{wang2025autonomous, Zhu2025groundeffect}. This necessitates control policies that can learn and adapt to these complex dynamics. Deep Reinforcement Learning (DRL) has emerged as a powerful paradigm for this, enabling agents to learn robust, end-to-end navigation policies directly from sensor data through trial-and-error interaction~\cite{kaufmann2023champion, drones8090516}.

\begin{figure}[t!]
    \centering
    \includegraphics[width=\columnwidth]{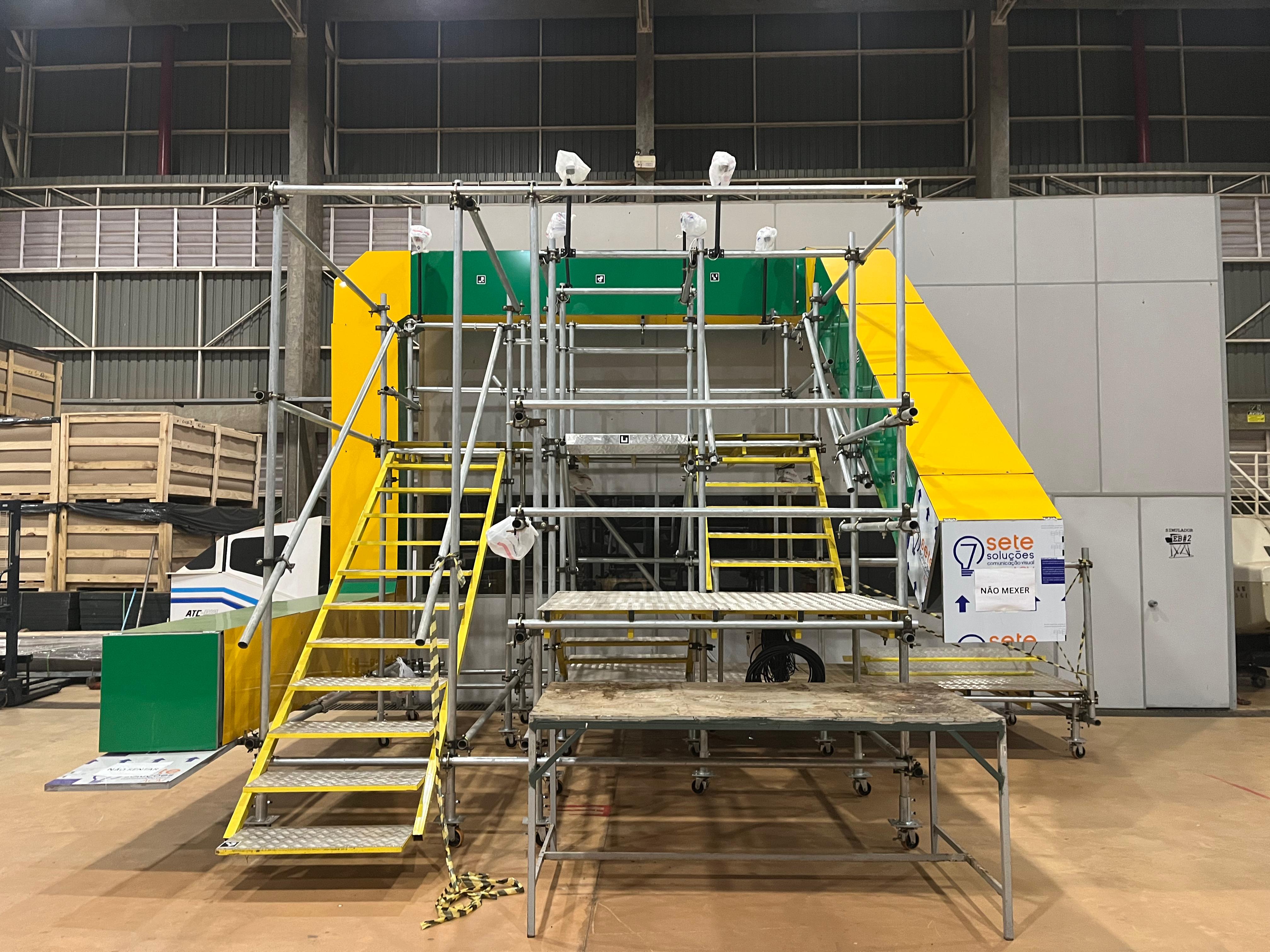}
    \\ 
    (a)
    \vspace{0.2cm}
    \\
    \includegraphics[width=\columnwidth]{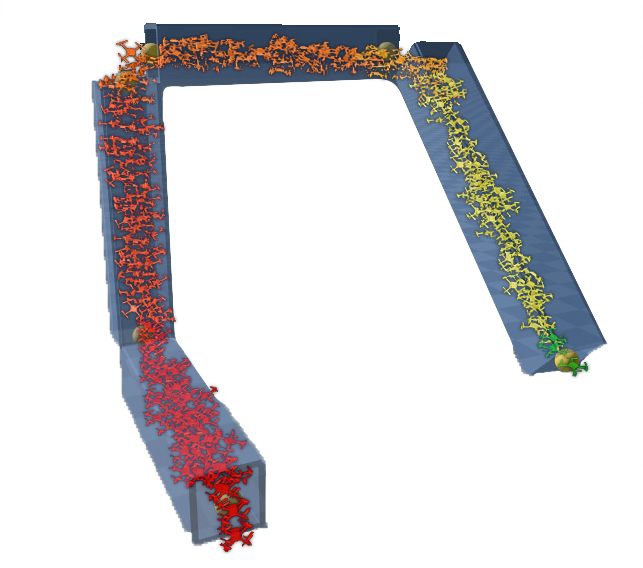}
    \\
    (b)
    \caption{(a) Experimental setup for drone navigation in confined spaces, including a duct, and (b) the drone navigation evolution during training when navigating inside the digital twin of the duct shown in (a). Red indicates unsuccessful navigation, which occurs when the drone fails at the beginning of the trajectory, and green indicates successful navigation.}
    \label{fig:ducts_combined}
\end{figure}

However, the choice of DRL algorithm is crucial, particularly for safety-critical applications such as duct inspection. A key distinction exists between off-policy and on-policy learning paradigms. Off-policy algorithms, such as Soft Actor-Critic (SAC)~\cite{pmlr-v80-haarnoja18b}, are highly sample-efficient as they reuse past experiences from a replay buffer. This trait is desirable for robotics, especially when considering fine-tuning on physical systems where real-world interactions are costly and potentially unsafe. In contrast, on-policy methods, such as Proximal Policy Optimization (PPO)~\cite{schulman2017proximal}, learn from freshly collected data, which often leads to more stable and reliable convergence, but at the cost of lower sample efficiency.

This trade-off motivates the central question of this work: for a task demanding high precision and safety, does the stability of on-policy methods outweigh the sample efficiency of off-policy algorithms? This study investigates this dichotomy by training and evaluating agents in a high-fidelity simulation, as illustrated in Fig.~\ref{fig:ducts_combined}, to determine the most suitable paradigm for reliable autonomous inspection.

The main contributions of this paper are threefold:
\begin{itemize}
    \item A direct comparative analysis of well-established on-policy and off-policy algorithms for the task of autonomous UAV navigation in confined industrial ducts.
    \item Empirical evidence demonstrating that for this hazard-dense, high-precision task, the training stability of the on-policy approach is more critical than the sample efficiency of the off-policy method, leading to a successful and robust policy.
    \item The validation of a simulation workflow using procedurally generated environments in a high-fidelity physics engine as a testbed for developing and benchmarking UAV control policies for industrial applications.
\end{itemize}

\section{Related Work}

DRL has revolutionized the development of autonomous systems by enabling agents to learn complex control policies through direct, trial-and-error interaction with their environment. Unlike traditional control methods that rely on precise analytical models, DRL allows an agent to discover optimal behaviors from high-dimensional sensor inputs, making it well-suited for robotics where real-world dynamics are often difficult or unfeasible to properly model~\cite{wu2023survey, alhamamid2022systematic}. For instance, foundational work in DRL has demonstrated its capability to generate dynamic and agile motor skills for legged robots, showcasing policies that can adapt to varied terrains and disturbances~\cite{Hutter2019, Tan2018}.

However, learning through real-world interaction is inherently challenging due to the high cost, slow pace, and potential safety risks associated with physical trial-and-error. Consequently, high-fidelity simulation has become an indispensable tool for DRL research, providing a safe, parallelizable, and cost-effective platform for policy training~\cite{song2020flightmare}. A central challenge in this paradigm is the gap between simulation and reality. Techniques such as domain randomization, which involves varying simulation parameters such as friction, lighting, and sensor noise, are critical for developing policies that are robust enough to transfer effectively to physical hardware~\cite{tobin2017domainrandomization, muratore2021simtoreal, kaufmann2020deep, crow_nav_2025}. Our work leverages the high-fidelity Genesis physics engine~\cite{Genesis} to create a realistic training environment that models the complex dynamics necessary for this task.

A prominent application of DRL in robotics is autonomous UAV navigation. The field has seen remarkable achievements, with DRL-trained policies demonstrating superhuman performance in high-speed, dynamic tasks such as drone racing~\cite{kaufmann2023champion, song2023reaching, hanover2024autonomous}. These successes prove that DRL can produce controllers that operate at the very limits of a system's physical capabilities. Although navigation in open spaces is a well-explored problem, operating in cluttered and confined environments, such as forests, urban canyons, or industrial ducts central to our work, presents a distinct and more severe set of challenges. In these settings, the proximity of surfaces induces complex, often unmodeled aerodynamic effects that significantly increase the risk of collision and destabilize flight~\cite{martin2025flying, wang2025autonomous, yang2025groundeffectawaremodelingcontrolmulticopters}.

Navigating these hazard-dense environments relies on the stability and reliability of the learning algorithm. Within DRL, a fundamental distinction exists between on-policy and off-policy methods. On-policy algorithms, such as PPO, learn exclusively from fresh data generated by the current policy and are often credited with greater training stability and more reliable policy convergence. Conversely, off-policy algorithms such as SAC utilize a replay buffer to learn from a vast history of past experiences, granting them superior sample efficiency, a highly attractive trait for robotics~\cite{learning2walk}. While both approaches are well-established~\cite{bakker2024benchmark}, the critical trade-off between on-policy stability and off-policy sample efficiency has not been thoroughly evaluated for precision navigation tasks in highly constrained settings. This paper addresses this gap by presenting a direct empirical comparison of PPO and SAC. The objective is to provide key insights into algorithm selection for robust, real-world robotic deployment where safety and reliability are paramount.

\definecolor{SACCol}{RGB}{240,240,240}
\definecolor{PPOCol}{RGB}{224,245,224}
\begin{table*}[t]
\caption{Definition of reward terms for the UAV navigation task (Section~III-C).}
\label{tab:drone_rewards_split}
\ieeetablesetup
\begin{tabular}{@{}llp{3.2cm}ccp{4.2cm}@{}}
\toprule
\multicolumn{2}{c}{\textbf{Reward Term}} & \textbf{Function} & \multicolumn{2}{c}{\textbf{Initial Weights}} & \textbf{Notes} \\
\cmidrule(lr){4-5}
 & & & \textbf{PPO} & \textbf{SAC} & \\
\midrule
\multirow{3}{*}{\textbf{Guidance}}
& Progress
& $(\mathbf{v}^{B}_{\mathrm{lin}} \cdot \Delta t)\,\dfrac{\mathbf{p}_{\mathrm{rel}}}{\|\mathbf{p}_{\mathrm{rel}}\|}$
& 25.0 & 50.0 & Rewards motion toward the next waypoint. \\
& Centerline Deviation
& $-\dfrac{\|p_t - c_t\|}{R_d}$
& 5.0 & 10.0 & Penalizes distance to duct centerline. \\
& Velocity Tracking
& $\exp\!\big(-\beta_v\,\|\mathbf{v}^{B}_{\mathrm{lin}} - v^\ast\|\big)$
& 3.0 & 4.0 & Encourages target forward speed $v^\ast$. \\
\midrule
\multirow{3}{*}{\textbf{Stability}}
& Orientation Alignment
& $\dfrac{\alpha_y f_B^\top d_h + \alpha_\ell u_B^\top u_W}{\alpha_y + \alpha_\ell}$
& 10.0 & 10.0 & Rewards yaw/level attitude. \\
& Angular Vel. Damping
& $-\|\mathbf{v}^{B}_{\mathrm{ang}}\|^2$
& $8.5\times10^{-3}$ & $8.5\times10^{-3}$ & Penalizes rotational speeds. \\
& Action Smoothness
& $-\|\mathbf{a}_t - \mathbf{a}_{t-1}\|^2$
& $7.0\times10^{-3}$ & $7.0\times10^{-3}$ & Penalizes abrupt motor changes. \\
\midrule
\multirow{3}{*}{\textbf{Event-based}}
& Waypoint Pass
& $\mathbb{I}\!\big[\|\mathbf{p}_{\mathrm{rel}}\| < 1.5\,R_d\big]$
& 22.0 & 22.0 & Sparse bonus near a waypoint. \\
& Duct Finish
& $\mathbb{I}\!\big[\text{all waypoints passed}\big]$
& 50.0 & 50.0 & Large terminal bonus for completion. \\
& Crash Penalty
& $-\mathbb{I}\!\big[\text{termination}\big]$
& 17.0 & 17.0 & Large terminal penalty (collision/violation). \\
\bottomrule
\end{tabular}

\vspace{0.25em}
\noindent\footnotesize \emph{Notes:} $R_d$ is the duct radius; $\mathbf{p}_{\mathrm{rel}}$ is the vector to the next waypoint; $f_B,u_B$ are body axes, $u_W$ world up; $\mathbb{I}[\cdot]$ is the indicator function.
\end{table*}


\section{Methods}

We present a full simulation workflow for DRL for one of the hazardous confined tasks in industrial inspection. We investigate two main consolidated algorithms, PPO and SAC, as representatives of on-policy and off-policy methods. All environments, problem framing, and reward engineering are carefully described in the following subsections.

\subsection{Problem Statement}
We pose goal-directed UAV control as a Markov Decision Process (MDP) $\mathcal{M}=(\mathcal{S},\mathcal{A},\mathcal{T},\mathcal{R},\gamma)$ with full observability from the simulator. $\mathcal{S}$ is the set of states $s_t$, $\mathcal{A}$ the set of actions $a_t$, $\mathcal{T}(s_{t+1}\!\mid\!s_t,a_t)$ the transition kernel induced by the physics engine and the actuation model in Eq.~\ref{eq:motor_control}, $\mathcal{R}(s_t,a_t)$ the reward, and $\gamma\in[0,1)$ the discount factor. The objective is to learn a policy $\pi^*$ that maximizes expected discounted return:
\begin{equation}\label{eq:mdp_objective}
\pi^* =\arg\max_{\pi}\ \mathbb{E}_{\pi,\mathcal{T}}\!\left[\sum_{k=0}^{\infty}\gamma^{k}\, \mathcal{R}(s_{t+k},a_{t+k})\right].
\end{equation}

The state vector $s_t\in\mathbb{R}^{20}$ aggregates geometric, kinematic, and actuation-history terms and is defined as
\begin{equation}\label{eq:state}
s_t=\big[\ {p}_{\mathrm{rel}},\ \hat{{p}}_{\mathrm{rel}}^{B},\ {q},\ {v}^{B}_{\mathrm{lin}},\ {v}^{B}_{\mathrm{ang}},\ {a}_{t-1}\ \big],
\end{equation}
where $\mathbf{p}_{\mathrm{rel}}\in\mathbb{R}^3$ is the position from the UAV to the next waypoint, $\hat{\mathbf{p}}_{\mathrm{rel}}^{B}\in\mathbb{R}^3$ its unit-normalized representation in body frame $B$, $\mathbf{q}\in\mathbb{R}^{4}$ is the unit quaternion (world-to-body), $\mathbf{v}^{B}_{\mathrm{lin}},\mathbf{v}^{B}_{\mathrm{ang}}\in\mathbb{R}^{3}$ are body-frame linear and angular velocities, and $\mathbf{a}_{t-1}\in\mathbb{R}^{4}$ is the previous motor-command vector. The continuous action $a_t\in[-1,1]^4$ parameterizes per-rotor commands; rotor speeds are obtained by
\begin{equation}\label{eq:motor_control}
\omega_i=\big(1+0.8\,a_{t,i}\big)\,\omega_{\mathrm{hover}},\quad i=1,\dots,4,
\end{equation}
with $\omega_{\mathrm{hover}} = 14.47~\mathrm{krpm}$ as the calibrated hover speed. Under this MDP, $\mathcal{S}$ and $\mathcal{A}$ are given by Eqs.~\ref{eq:state}–\ref{eq:motor_control}, $\mathcal{T}$ is the stochastic dynamics induced by the simulator and actuation mapping, and $\mathcal{R}$ encodes waypoint-tracking performance with regularization on motion and control effort (dense shaping compatible with Eq.~\ref{eq:mdp_objective}).

\subsection{Simulation Environment}
The experiments were conducted in the Genesis~\cite{Genesis} physics engine, a platform enabling parallel, GPU-accelerated rigid-body simulation. The training environment features a procedurally generated duct for each episode, ensuring the policy learns to navigate diverse and challenging scenarios.

\begin{figure}[ht]
    \centering
    \includegraphics[width=0.8\columnwidth]{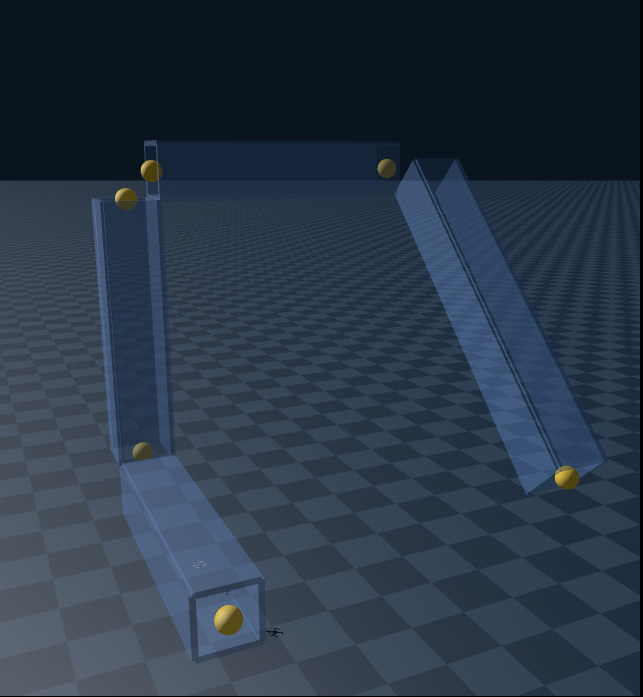}
    \caption{Procedurally generated duct environment in Genesis, with a connected tubular segments with varying orientations.}
    \label{fig:tubes}
\end{figure}

Each duct is constructed as a sequence of $N_s$ straight tubular segments connected end-to-end, as depicted in Fig.~\ref{fig:tubes}. While the sections of the squared tubes share a side of $R_d = 0.5m\,\mathrm{m}$ and have stochastically determined lengths, their orientation is the key to creating complex paths. The angular deviation between consecutive tubes is randomized using Rodrigues' rotation formula~\cite{kan2019analysis}, given by Eq.~\ref{eq:rodrigues}. Specifically, the direction vector of a new segment is computed by applying a controlled rotation to the vector of the previous segment. This method ensures a continuous and natural change in the duct's path, emulating the structure of real-world pipelines.
\begin{equation}\label{eq:rodrigues}
{v}' = {v}\cos\theta + ({k} \times {v})\sin\theta + {k}({k} \cdot {v})(1 - \cos\theta)
\end{equation}

\(\mathbf{v}\) is the vector to rotate, \(\mathbf{k}\) is the unit vector along the axis of rotation (the plane normal), and \(\boldsymbol{\theta}\) is the rotation angle ($\theta \in (-90.0^{\circ},90.0^{\circ})$). Then, the generator outputs the walls and waypoints for each segment, which are then utilized in the simulation for both navigation and collision testing.

\begin{figure}[ht]
    \centering
    \includegraphics[width=0.35\textwidth]{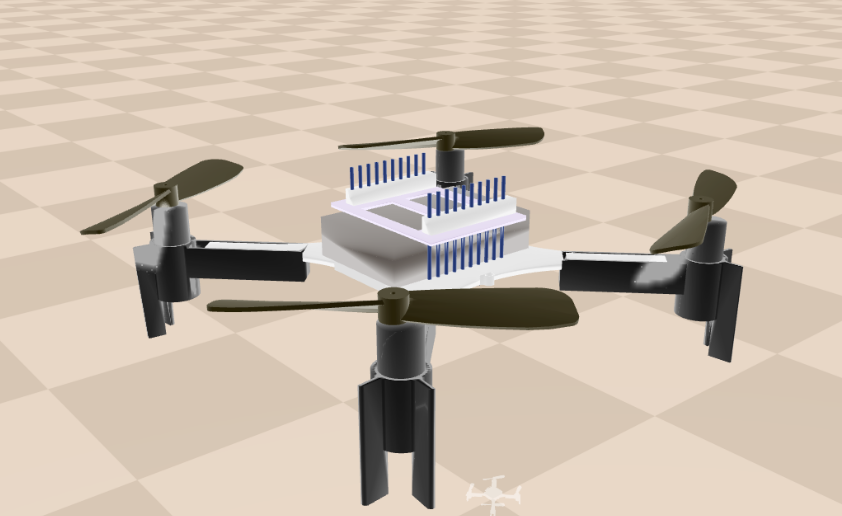}
    \caption{Simulated model used for all experiments of a Crazyflie 2, a $92 \times 92 \times 29$~mm nano-quadcopter.}
    \label{fig:crazyflie}
\end{figure}

The waypoints $\{\mathbf{w}_i\}_{i=1}^{N_w}$ are placed along the duct's centerline to guide the agent through these generated courses. The agent controls a simulated Bitcraze Crazyflie 2 (CF2), a $92 \times 92 \times 29$~mm nano-quadcopter (Fig.~\ref{fig:crazyflie}), although the framework supports any UAV via URDF import.

\subsection{Learning Algorithms}

We compare on-policy and off-policy paradigms using a single framework, \texttt{skrl}~\cite{serrano2023skrl}, to eliminate cross-library discrepancies and strengthen procedural rigor and reproducibility. Both agents operate on the MDP specified above and share the same network backbone: an actor–critic with two hidden layers ($256,128$ units, ELU). For the on-policy case, PPO in \texttt{skrl} uses a rollout horizon of $256$ with $4096$ parallel environments, an adaptive KL target of $0.01$, discount $\gamma=0.99$, GAE parameter $\lambda=0.95$, and clipping parameter $\epsilon=0.2$. For the off-policy case, SAC in \texttt{skrl} employs an actor and twin critics with the same architecture, a replay buffer of $10^6$ transitions, batch size $512$, target update $\tau=0.005$, discount $\gamma=0.99$, automatic entropy tuning, and a learning rate linearly decayed from $3\times 10^{-2}$ to $0$ over $10^6$ timesteps. Consolidating PPO and SAC within \texttt{skrl} ensures an identical data path (vectorized rollouts and replay), optimizer/scheduler behavior, logging, and device placement, thereby isolating algorithmic effects from implementation artifacts and enabling fair, repeatable comparisons.

\subsection{Reward Formulation}
\label{sec:reward_formulation}

The reward function is a critical component for guiding the agent's learning process. To address the multi-faceted nature of the UAV navigation task, a composite reward function, $R_t$, was designed and computed at each timestep as a weighted sum of several terms:
\begin{equation}
    R_t = \sum_{k} w_k\, r_k,
    \label{eq:reward_sum}
\end{equation}

where $r_k$ represents an individual reward component and $w_k \in \mathbb{R}^{+}$ is its corresponding weight, detailed in Table~\ref{tab:drone_rewards_split}. 

This modular structure is common in robotics, as it allows for the decomposition of a complex task into more manageable sub-objectives \cite{Hutter2019, Tan2018}. The formulation is divided into three distinct categories. \textbf{Guidance rewards} provide a dense signal to direct the agent along the desired path by rewarding forward \textit{Progress}, penalizing \textit{Centerline Deviation}, and encouraging \textit{Velocity Tracking} \cite{kaufmann2020deep}. Complementing this, \textbf{Stability rewards} act as a regularization to promote physically safe flight, constraining \textit{how} the task is performed. These include terms for maintaining proper \textit{Orientation Alignment}, damping high \textit{Angular Velocity}, and ensuring \textit{Action Smoothness} to prevent abrupt motor commands, which is a standard practice for robust control in locomotion tasks \cite{Tan2018, Hutter2019}. Finally, \textbf{Event-based rewards} deliver sparse, high-magnitude signals for critical discrete events, such as a bonus for a \textit{Waypoint Pass}, a large terminal reward for \textit{Duct Finish}, or a penalty for a \textit{Crash} \cite{Hutter2019}.


To determine the weights $w_k$ and optimize the training process, a central challenge in deep reinforcement learning \cite{li2018hyperband}, we utilized the sweep tool from the Weights \& Biases platform. This method explored a predefined range of values for each reward parameter, replacing empirical adjustments with an efficient search for the optimal configuration. The objective was to identify the combination of weights that maximized the average reward value of the PPO and SAC agents, finding the ideal configuration for each algorithm. It is known that off-policy algorithms such as SAC, which learn from a diverse replay buffer, can exhibit different sensitivities to reward scaling compared to on-policy methods \cite{Haarnoja2018}. A hypothesis was formed that a stronger signal for the primary task objectives could improve learning from the varied experiences in the buffer. Consequently, the range of weights set for the main guidance terms (\textit{Progress} and \textit{Centerline Deviation}) were increased for SAC. Despite this optimization, the agent's performance remained suboptimal, as discussed in Section~\ref{sec:results}. This work narrows its scope to testing both algorithms with the aim of optimizing their performance. We recognize, however, that a more comprehensive ablative analysis can be done to improve these claims. This limitation will be addressed in future research through a more detailed investigation.



\section{Results}\label{sec:results}


\begin{table*}[t]
\caption{PPO performance metrics for selected training checkpoints.}
\label{tab:performance_evolution}
\ieeetablesetup
\begin{tabular}{@{}lcccccccc@{}}
\toprule
\textbf{Checkpoint} & \textbf{50} & \textbf{75} & \textbf{100} & \textbf{150} & \textbf{200} & \textbf{300} & \textbf{400} & \textbf{500} \\
\midrule
Average Reward            & 1.3k & 2.7k & 4.5k & 6.4k & 7.2k & 9.9k & \textbf{10.2k} & 9.6k \\
Avg. Waypoints Passed     & 1/7  & 2/7  & 4/7   & 5/7  & 6/7  & 7/7  & 7/7            & 7/7  \\
Avg. Collisions / Episode & 1.00 & 0.70 & 0.30 & 0.00 & 0.00 & 0.00 & 0.00           & 0.00 \\
Average Deviation (m)     & 0.123 & 0.113 & 0.084 & 0.065 & 0.094 & 0.064 & 0.063 & 0.094 \\
Maximum Deviation (m)     & 0.232 & 0.219 & 0.151 & 0.182 & 0.239 & 0.200 & 0.195 & 0.239 \\
\bottomrule
\end{tabular}
\end{table*}

\subsection{PPO Training}
The training progression of the PPO agent, detailed in Table~\ref{tab:performance_evolution}, demonstrates a convergence to a robust and effective navigation policy. Early in the training, by the first checkpoint (300 iterations), the agent had already achieved a 100\% course completion rate with zero collisions. This indicates that the fundamental task of traversing the duct without catastrophic failure was learned quickly, validating the efficacy of the reward function's core guidance and penalty terms.

\begin{figure}[!ht]
    \centering
        \includegraphics[width=\columnwidth]{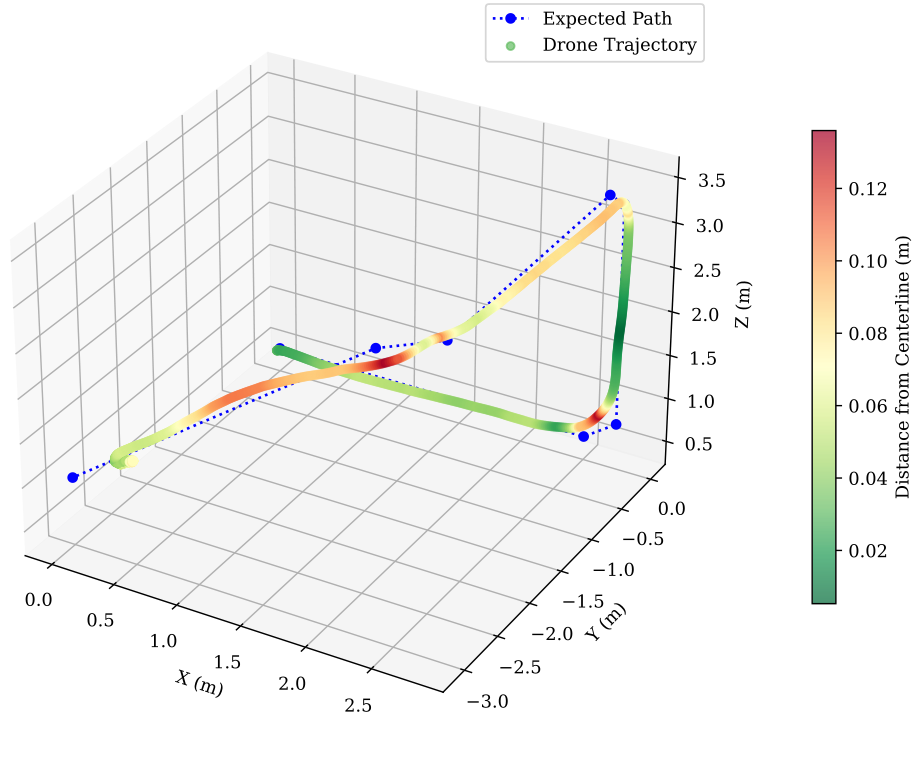}
    \caption{The drone trajectory that achieves the best performance for a model
trained with PPO algorithm after 400 checkpoints.}
    \label{fig:trajectory_comparison}
\end{figure}

The subsequent training phase focused on refining the agent's control policy for improved precision and stability (Fig.~\ref{fig:trajectory_comparison}). A significant enhancement in flight quality is observed between checkpoints 200 and 300, where the {Average Deviation} from the centerline was nearly halved, decreasing from 0.1128~m to 0.0636~m. This period corresponds to the agent learning to minimize lateral drift and maintain a more centered trajectory, a critical behavior for navigating narrow passages. The policy continued to optimize, reaching its peak performance at checkpoint 400, where it recorded the highest {Average Reward} (10.2k) and maintained a low average deviation. The slight performance degradation at the final checkpoint (500) may suggest minor overfitting or the policy settling into a slightly different but equally successful equilibrium. Overall, the PPO agent demonstrated a classic learning pattern: mastering the primary objective first, followed by a clear phase of trajectory optimization. The resulting policy was more stable and precise than that of SAC (Subsection~\ref{sactrain}) and successfully completed the entire proposed task.


\begin{table*}[t]
\caption{SAC performance metrics for selected training checkpoints.}
\label{tab:my_performance_data}
\ieeetablesetup
\begin{tabular}{@{}lcccccccc@{}}
\toprule
\textbf{Checkpoint} & \textbf{50} & \textbf{75} & \textbf{100} & \textbf{150} & \textbf{200} & \textbf{300} & \textbf{400} & \textbf{500} \\
\midrule
Average Reward            & 2.0k & 3.0k & 3.6k & 4.1k & \textbf{5.4k} & 4.4k   & —   & —   \\
Avg. Waypoints Passed     & 0/7  & 1/7  & 2/7  & 3/7  & 3/7           & 3/7   & —   & —   \\
Avg. Collisions / Episode & 1.00 & 1.00 & 1.00 & 1.00 & 1.00          & 1.00   & —   & —   \\
Average Deviation (m)     & 0.037 & 0.049 & 0.065 & 0.156 & 0.058      & 0.61   & —   & —   \\
Maximum Deviation (m)     & 0.075 & 0.117 & 0.192 & 0.199 & 0.112      & 0.134   & —   & —   \\
\bottomrule
\end{tabular}
\end{table*}

\subsection{SAC Training}
\label{sactrain}
In contrast to the PPO agent, the SAC agent failed to develop a successful end-to-end navigation policy, as evidenced by the performance metrics in Table~\ref{tab:my_performance_data}. Throughout the entire training process of 650k timesteps, the {Course Finish Rate} remained at 0.0\%, with the agent consistently failing to traverse the entire duct system. Collisions were persistent, with an average of 1.00 per episode, indicating that terminal failure was the standard outcome.

\begin{figure}[ht!]
    \centering
    \includegraphics[width=0.5\textwidth]{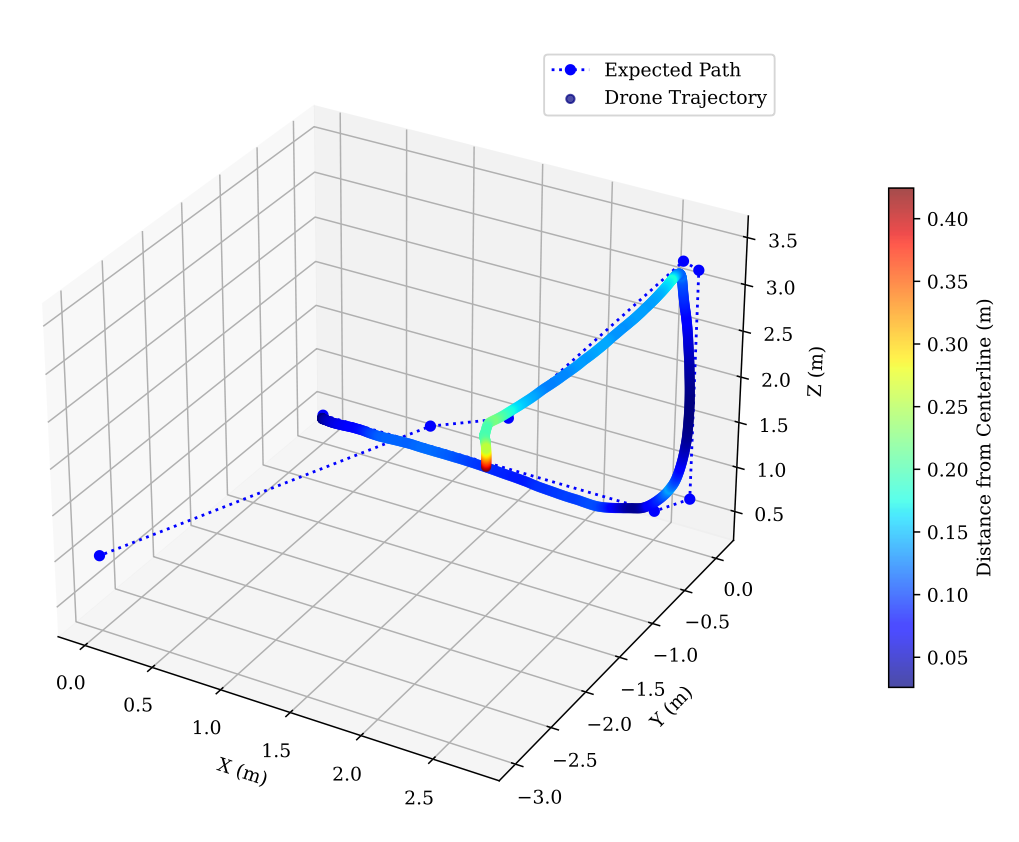}
    \caption{The drone trajectory that achieves the best performance for a model trained with SAC algorithm after 200 checkpoints.}
    \label{fig:sac_trajectory_530k}
\end{figure}

Despite this overarching failure, the data reveals that the agent did engage in a learning process. The {Average Reward} shows a consistent upward trend, increasing from 2.0k to 5.4k. This improvement is correlated with the agent's ability to navigate a limited portion of the course, successfully passing a maximum of three waypoints before crashing. This behaviour is characteristic of an off-policy agent converging to a local optimum. The replay buffer, heavily populated with experiences from the initial, simpler segments of the duct, likely biased the agent towards perfecting the start of the trajectory at the expense of discovering strategies to overcome later challenges. The sample efficiency of SAC, therefore, proved counterproductive in this context, as it led the agent to over-specialize on a suboptimal, incomplete behaviour pattern without the broader (Fig.~\ref{fig:sac_trajectory_530k}), on-policy exploration needed to find a globally successful solution.

\section{Conclusion}

This work investigated the application of Deep Reinforcement Learning for autonomous UAV navigation in confined industrial ducts, a task where precision and safety are critical. We framed the problem as a direct comparison between on-policy (PPO) and off-policy (SAC) paradigms to determine the most suitable approach for such a high-stakes environment. Our results provide empirical evidence that for this hazard-dense, high-precision task, the training stability afforded by an on-policy algorithm is more critical than the sample efficiency of an off-policy method. The PPO agent successfully learned a robust, collision-free policy capable of completing the entire course, whereas the SAC agent consistently failed to find a complete solution, converging instead to a local optimum.

The divergence in performance highlights a key challenge for DRL in safety-critical robotics. We attribute SAC's failure to its replay buffer, which became saturated with experiences from the initial, simpler segments of the duct. This biased the learning process, preventing the agent from exploring and mastering the more challenging, distal parts of the trajectory. In contrast, PPO's on-policy updates from fresh interaction data provided a more consistent learning signal, enabling it to overcome local optima and solve the end-to-end task. This finding suggests that for complex sequential tasks where reliable convergence to a safe policy is paramount, the stability of on-policy learning can be the decisive factor, outweighing the theoretical benefits of off-policy sample efficiency.

Having validated this methodology in a high-fidelity simulation, future work will focus on bridging the sim-to-real gap. The primary objective is to transfer the successful PPO policy to a physical UAV testbed. This transfer will be facilitated by incorporating domain randomization and curriculum learning into the training regimen to ensure the policy is robust to real-world uncertainties. Furthermore, we plan to investigate advanced off-policy or hybrid algorithms designed to mitigate the exploration challenges observed here, aiming to unify the stability of on-policy methods with the data efficiency required for practical robotic applications.







\bibliographystyle{IEEEtran}
\bibliography{ref}

\end{document}